\pdfoutput=1
\documentclass{article}
\usepackage{spconf,amsmath,graphicx,amssymb}

\usepackage{multirow}
\usepackage{graphicx}
\usepackage{kotex}
\usepackage{amsmath}

\usepackage{adjustbox}
\usepackage{subcaption}
\usepackage{multirow}
\usepackage{microtype}
\usepackage{nicefrac}
\usepackage{booktabs}
\usepackage{amsfonts} 
\usepackage{color}
\usepackage{url}
\usepackage{tabularx}

\usepackage{adjustbox}

\usepackage{bm}
\newcommand{\bs}[1] {\bm{#1}}

\usepackage{algorithm}
\usepackage{algpseudocode}
\usepackage[numbers]{natbib}

\makeatletter
\def\hlinewd#1{%
	\noalign{\ifnum0=`}\fi\hrule \@height #1 %
	\futurelet\reserved@a\@xhline}
\makeatother

\title{Learning Imbalanced Datasets with Maximum Margin Loss}
\name{Haeyong Kang$^{\ast}$, \qquad Thang Vu, \qquad Chang D. Yoo}
\address{Korea Advanced Institute of Science and Technology\\
	School of Electrical Engineering\\
	291 Daehak-ro, Yuseong-gu, Daejeon, 34141, Republic of Korea
}
\begin{document}
	%
	\maketitle
	\begin{abstract}
		A learning algorithm referred to as Maximum Margin (MM) is proposed for considering the class-imbalance data learning issue:the trained model tends to predict the majority classes rather than the minority ones. That is, underfitting for minority classes seems to be one of the challenges of generalization. For a good generalization on the minority classes, we design a new Maximum Margin (MM) loss function, motivated by minimizing a margin-based generalization bound through the shifting decision bound. The theoretically-principled label-distribution-aware margin (LDAM) loss was successfully applied with prior strategies such as re-weighting or re-sampling, along with the effective training schedule. However, they didn't investigate the maximum margin loss function yet. In this study, we investigate the performances of two types of hard maximum margin based decision boundary shift with LDAM's training schedule on artificially imbalanced CIFAR-10/100 for fair comparisons and show the effectiveness. Code is available at \url{https://github.com/ihaeyong/Maximum-Margin-LDAM}.
	\end{abstract}

	\section{Introduction}
\begin{figure}
		\centering
		\includegraphics[width=1.0\linewidth,trim={0. 0. 0. 0.},clip]{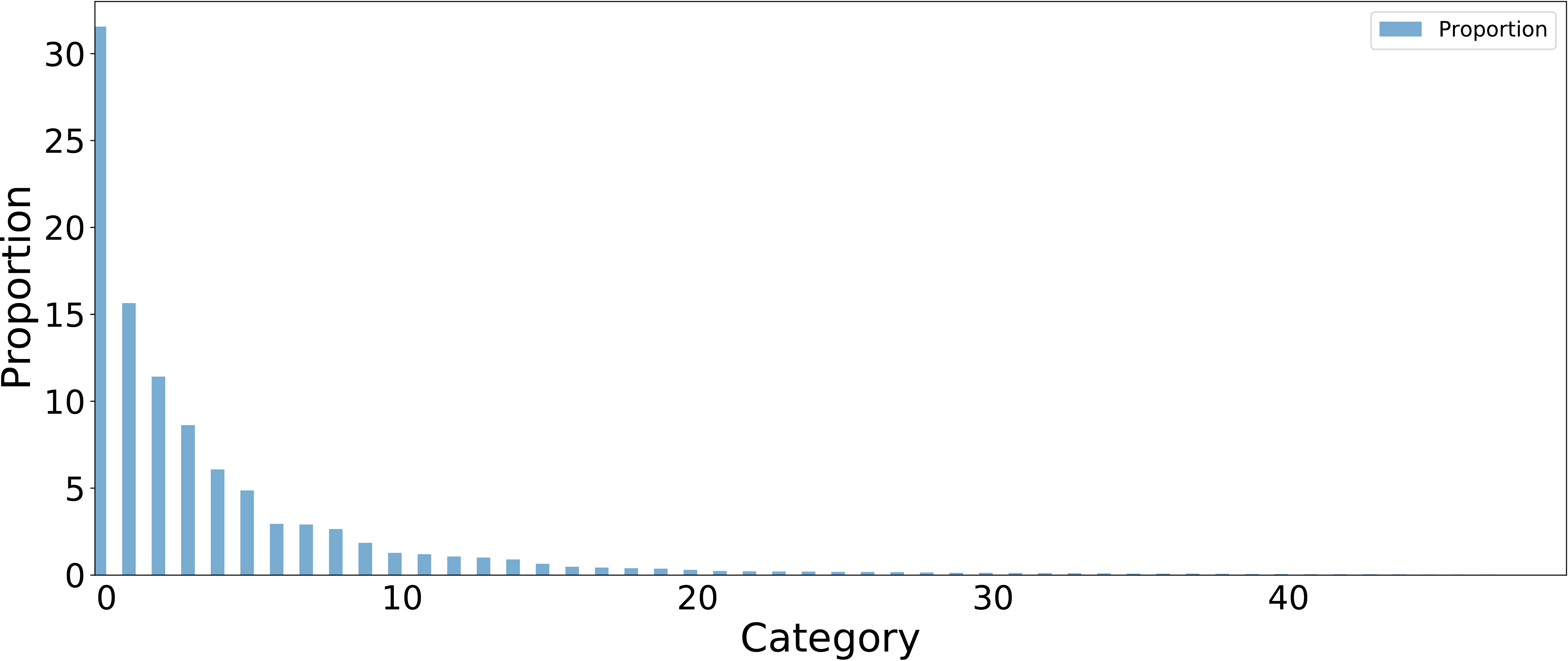}
		\caption{Long-tail distribution of the real-world large datasets \cite{krishna2017visual}. Under these extremely imbalanced datasets, the deep model suffers from over-fitting on the minority classes.}
		\label{fig:long_tail}
	\end{figure}

	With the advancement of the deep neural networks, large scale datasets have appeared. In general, these real-world large data sets often have shown long-tailed label distributions \cite{van2017devil, lin2014microsoft, liu2019large} as shown in Fig.\ref{fig:long_tail}. On these datasets, the models have been shown to perform poorly on the minority classes, over-fitting to the minority classes. This factor has to do with biased predictions. For example, the trained model tends to predict the majority classes rather than the minroity ones as shown in Fig 1. Overfitting for majority classes seems to be one of the challenges of generalization.
	
	For robustness to the over-fitting to minority classes, it needs to design a training loss that is in expectation closer to the test distribution or to regularize the parameters to achieve better trade-offs between the accuracies of the majority classes and the minority classes. Instead of depending on the sampling size-dependent margins \cite{cao2019learning}, we design a hard maximum margin loss function that encourages the model to have the optimal trade-off between per-class margins. 
	
\begin{figure}
		\centering
		\includegraphics[width=1.0\linewidth,trim={5. 5. 5. 5.},clip]{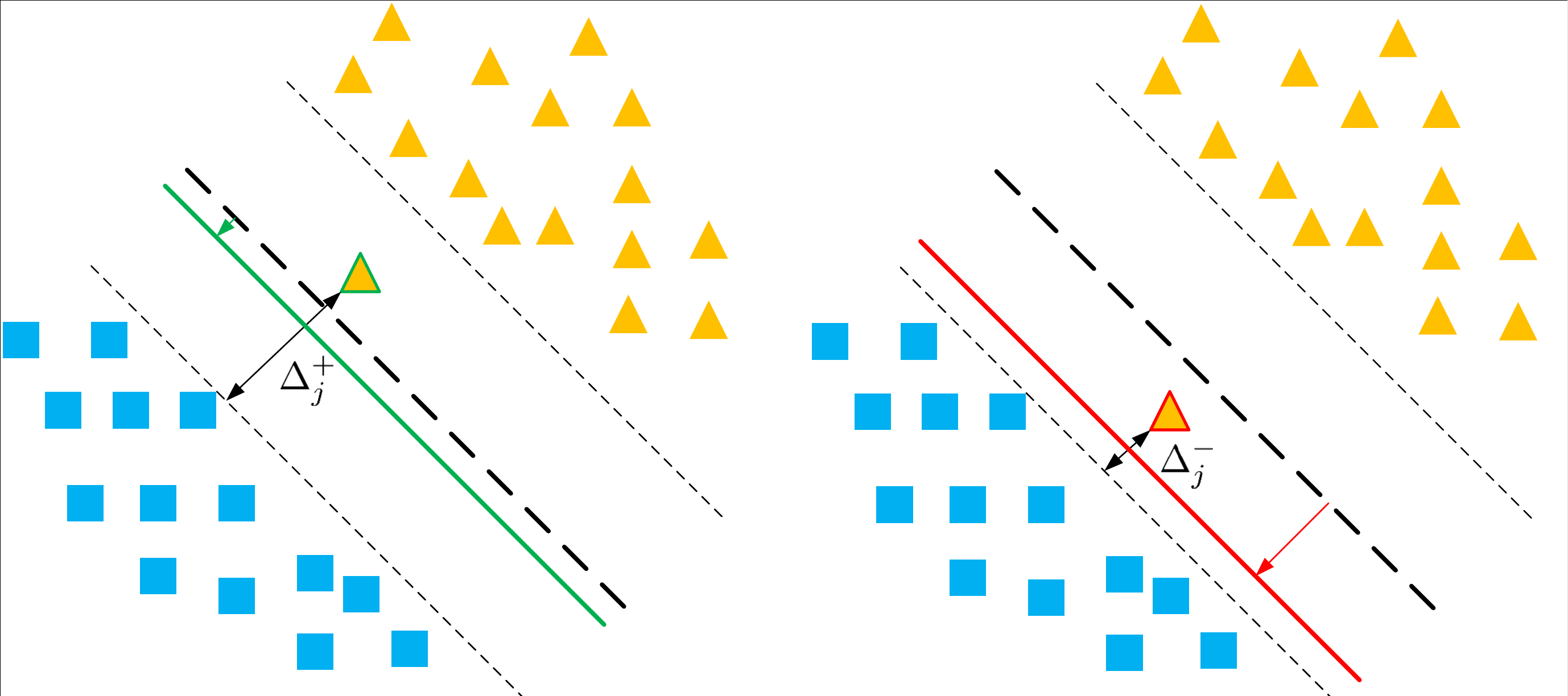}
		\caption{For classification with a linearly separable classifier, the maximum margin of the $j$-th class $\Delta_j$ is defined to be the minimum distance of the data in the $j$-th class to the decision boundary (dotted line). As illustrated here, we assume that the decision boundaries are shifted by two types of the hard maximum margin of samples: hard positive margin $\Delta_j^{+}$ and hard negative margin $\Delta_j^{-}$ respectively;  our loss function device $\Delta_j^{-}$ to occupy more margin than $\Delta_j^{+}$ such that the red decision boundary shifts more than the green one.}
		\label{fig:outline}
	\end{figure}

	To achieve an optimal trade-off between the margins of the classes, we design a loss function to maximize the per-class margins with the following assumption. See figure \ref{fig:outline} for an illustration in the binary classification case. We assume that the decision boundary is shifted by the hard samples that are defined by the maximum margin. The hard samples compose of two types of margins: hard positive margin $\Delta_j^{+}$ and hard negative margin $\Delta_j^{-}$. The hard positive margins $\Delta_j^{+}$ w.r.t $j$-th class are defined by the maximum margin with correctly classified samples; the hard negative margins $\Delta_j^{-}$ w.r.t $j$-th class are defined by the maximum margin with miss-classified samples. In the training, the hard negative margins shift the model's decision boundary more than the positive margins.
	
	In summary, our main contributions are (1) we design a maximum margin loss function to encourage larger sample margins for hard negative sample classes such that the smaller the maximum margins are the greater the shifting margins are. (2) we applied the maximum loss to the deferred re-balancing optimization procedure \cite{cao2019learning} for more generalization, and (3) our practical implementation shows significant improvements on two benchmark vision tasks, such as artificially imbalanced CIFAR-10/100 for fair comparisons.
	
	This paper is organized as follows. Section \ref{sec:related_work} provides discussions of related works on imbalanced learning and maximum margin loss function. In Section \ref{sec:hmm_learn}, the proposed maximum margin loss function is discussed. In Section \ref{sec:exper}, experimental results on the artificially imbalanced CIFAR-10/100 are discussed along with ablation tests. Finally, Section \ref{sec:conc} concludes.
	
	\section{Related Works} \label{sec:related_work}
	The two classical approaches for learning long-tailed data are re-weighting \cite{cui2019class} the losses of the examples and re-sampling (over-sampling the minority classes \cite{byrd2019effect} and under-sampling the majority classes \cite{buda2018systematic}) the examples in the SGD mini-batch. They both devise a training loss that is in expectation closer to the test distribution to achieve better trade-offs between the accuracies of the majority classes and the minority classes. Recently, other learning paradigms have also been explored such as transfer learning \cite{liu2019large}, metric learning \cite{you2018scalable}, meta-learning \cite{shu2019meta}, semi-supervised and self-supervised learning \cite{yang2020rethinking}, and decoupled representation and classifier \cite{zhou2020bbn, kang2019decoupling}.
	
	
	
	A maximum-margin classifier is typically obtained by using the hinge loss function in SVMs \cite{suykens1999least}. The maximum-margin classifier benefits from margins to minimize intra-class variation in predictions and to maximize the inter-class margin. With the benefits of the maximum-margin, Large-Margin Softmax \cite{liu2016large}, Angular Softmax \cite{liu2017sphereface}, and Additive Margin Softmax \cite{wang2018additive} have been proposed recently. In contrast to these class-independent margins, Label-Distribution-Aware Margin (LDAM) encourages bigger margins for minority classes, providing a concrete formula for the desired margins of the classes. Uneven margins for imbalanced datasets are also proposed and studied in \cite{li2002perceptron, khan2019striking}. However, they didn't investigate the maximum margin loss led by training samples yet. In this study, we investigate the performances of two types of hard maximum margin based decision boundary shift, comparing the results with current state-of-the-art methods.

	\section{Maximum Margin (MM) Learning}\label{sec:hmm_learn}
		
	\subsection{Maximum Margin (MM) Loss}
	
	Inspired by the trade-off between the class margins for two classes, we define two types of maximum margins for multiple classes of the following form as follows :
	
	\begin{equation}
	\Delta_y^{MM} = \begin{cases}
	\Delta_y^{+} \;\;\; \text{  if  }  f(x) == y; \\
	\Delta_y^{-} \;\;\; \text{  otherwise.}
	\end{cases}
	\label{eq:hmm}
	\end{equation}
	
	\noindent 
	where an example $(x, y)$, a deep model $f$ with logits $\bs{z}$, hyper-parameters $\delta^{+}/\delta^{-}$ to acquire empirical sample maximum margins, 
	\begin{eqnarray}
	\Delta^{+}_y = \exp\left(-\max(z_y - \max_{j \neq y} {z_j}, 0) - \delta^{+} \right),
	\label{eq:hmm_pos}
	\end{eqnarray} and 	
	\begin{eqnarray}
	\Delta^{-}_y = \exp\left(-\max(\max_{j \neq y} {z_j} - z_y , 0) - \delta^{-} \right).
	\label{eq:hmm_neg}
	\end{eqnarray}

	To achieve an optimal trade-off between the margins of the classes, we design a Maximum Margin (MM) loss function to maximize the per-class margins with the following assumption. See figure \ref{fig:outline} for an illustration in the binary classification case. We assume that the decision boundary is shifted by the hard samples that are defined by the maximum margin. The hard samples compose of two types of margins: hard positive margin $\Delta_j^{+}$ and hard negative margin $\Delta_j^{-}$ with hyper-parameter $\delta$ for smoothing effects. The hard positive margins $\Delta_j^{+}$ w.r.t $j$-th class are defined by the maximum margin with correctly classified samples ($z_y > \max_{j \neq y} {z_j}$). The hard negative margins $\Delta_j^{-}$ w.r.t $j$-th class are defined by the maximum margin with miss-classified samples ($ \max_{j \neq y} {z_j} > z_y$). We take exponential function to get more non-linearity such that the smaller the maximum margins are the greater the shifting margins are.
	
	We design a maximum margin loss function to encourage the network to have the margins above. Let $(x, y)$ be an example and $f$ be a model. For simplicity, we use $z_j = f(x)_j$ to denote the $j$th-output of the model for the $j$-th class. Following the previous work \cite{cao2019learning}, in order to tune the margin more easily, we effectively normalize the logits (the input to the loss function) by normalizing the last hidden activation to $\ell_2$ norm $1$ and normalizing the weight vectors of the last fully-connected layer to $\ell_2$ norm $1$. Notice that we then scale the logits by a constant $s = 10$. Empirically, the non-smoothness of hinge loss may pose difficulties for optimization. The smooth relaxation of the hinge loss is the following cross-entropy loss with enforced margins:
	
	\begin{equation}
	\mathcal{L}_{MM} ((x,y); f) = -\log \frac{e^{z_y - \Delta_y^{MM}}}{e^{z_y - \Delta_y^{MM}} + \sum_{j\neq y} e^{z_j}}
	\end{equation}
	where $\Delta_j^{MM}$ in Eq. \ref{eq:hmm} for $j \in \{1, \cdots, k\}$.

	\subsection{MM's Hyper-parameters $\delta^{+}/\delta^{-}$}
	To achieve the best performances of MM, we set the relationship between the hyper-parameters $\delta^{+}, \delta^{-}$ in Eq. \ref{eq:hmm_pos} and \ref{eq:hmm_neg} as follows:
	\begin{equation}
	\delta^{+} =  \delta^{-} * \beta
	\label{eq:delta}
	\end{equation}
	where the $\beta$ is a scaler. In the experiments, the best performances were acquired by setting both $\delta^{-}$ and $\beta > 1.0$ empirically, meaning that the hard negative margins $\Delta_j^{-}$ shift more than $\Delta_j^{+}$ in the process of training. To further enforce a class-dependent margin for multiple classes, we add the class-distribution-aware margin $\gamma_j = \frac{C}{n_j^{1/4}}$ \cite{cao2019learning} for some constant $C$ to Eq. \ref{eq:delta} as follows:  
	\begin{eqnarray}
	\delta^{+}_j = (\delta^{-} - \gamma_j) * \beta,
	\label{eq:delta_pos}
	\end{eqnarray} and 	
	\begin{eqnarray}
	\delta^{-}_j = \delta^{-} - \gamma_j.
	\label{eq:delta_neg}
	\end{eqnarray}

	\subsection{Deferred Re-balancing Optimization Schedule \cite{cao2019learning}}
	For a fair comparison, the proposed MM loss is also applied to the deferred re-balancing training procedure \cite{cao2019learning} as shown in Algorithm \ref{al:effective}, which first trains using vanilla ERM with the MM loss before annealing the learning rate, and then deploys a re-weighted MM loss with a smaller learning rate. In the following experiments with the MM loss function, the first stage of training leads to better initialization for the second stage of training with re-weighted losses. With the non-linear MM loss of the hyper-parameter $\delta^{+},\delta^{-}$ and deferred re-balancing training, the re-weighting scheme works stable more. 	
	\begin{algorithm}
		\caption{Imbalanced Learning with MM Loss}
		\label{al:effective} 
		\begin{algorithmic}[1]
			\Require Dataset $\mathcal{D} = \{(x_i, y_i)\}_{i=1}^n$, A model $f_\theta$
			\State Initialize the model parameters $\theta$ randomly
			\For {$t=T_0,T_1,,\ldots,T_0$}
			\State $\mathcal{B} \leftarrow$ SampleMinibatch($\mathcal{D}$, $m$)
			\State $\mathcal{L}(f_\theta) \leftarrow \frac{1}{m} \sum_{(x,y) \in \mathcal{B}} \cdot \mathcal{L}_{MM}((x,y); f_\theta)$
			\State $f_\theta \leftarrow f_\theta - \alpha \nabla_\theta \mathcal{L}(f_\theta)$
			\EndFor
			\For {$t=T_0,,\ldots,T$}
			\State $\mathcal{B} \leftarrow$ SampleMinibatch($\mathcal{D}$, $m$)
			\State $\mathcal{L}(f_\theta) \leftarrow \frac{1}{m} \sum_{(x,y) \in \mathcal{B}}  n_y^{-1}\cdot \mathcal{L}_{MM}((x,y); f_\theta)$
			\State $f_\theta \leftarrow f_\theta - \alpha \nabla_\theta \mathcal{L}(f_\theta)$
			\EndFor
		\end{algorithmic} 
	\end{algorithm}

	\section{Experiments} \label{sec:exper}
	\textbf{Datasets.} We evaluate our proposed MM loss function on artificially created versions of CIFAR-10 and CIFAR-100 with controllable degrees of data imbalance.
	
	\begin{table*}[t]
			\center
			\caption{Top-1 validation errors of ResNet-32 on imbalanced CIFAR-10 and CIFAR-100. The MM-LDAM-DRW, achieves better performances, and each of them individually is beneficial when combined with LDAM loss or DRW schedules.}
			\tiny
			\begin{adjustbox}{width=.75\textwidth}
			\begin{tabular}{l|ll|ll|ll|ll}
				\hlinewd{1.1pt}
				Dataset         & \multicolumn{4}{l|}{Imbalanced CIFAR-10}                            & \multicolumn{4}{l}{Imbalanced CIFAR-100}                           \\ \hline
				Imbalance Type  & \multicolumn{2}{l|}{long-tailed} & \multicolumn{2}{l|}{step}        & \multicolumn{2}{l|}{long-tailed} & \multicolumn{2}{l}{step}        \\ \hline
				Imbalance Ratio & \multicolumn{1}{l|}{100} & 10    & \multicolumn{1}{l|}{100} & 10    & \multicolumn{1}{l|}{100} & 10    & \multicolumn{1}{l|}{100} & 10    \\ \hline
				ERM   \cite{cao2019learning} & 29.64  & 13.61 & 36.70 & 17.50 & 61.68 & 44.30 & 61.45 & 45.37 \\
				Focal \cite{lin2017focal}& 29.62 & 13.34 & 36.09 & 16.36 & 61.59 & 44.22 & 61.43 & 46.54 \\
				LDAM  \cite{cao2019learning} & 26.65  & 13.04 & 33.42 & 15.00 & 60.40 & 43.09 & 60.42 & 43.73 \\ \hline 
				\textbf{MM} (ours) & \textbf{26.56} & \textbf{12.34}&  \textbf{33.19}  &\textbf{13.99}& \textbf{60.29} & \textbf{42.63} &\textbf{60.25}  & \textbf{43.55}  \\ \hline \hline
				CB RS \cite{cao2019learning} & 29.45  & 13.21 & 38.14 & 15.41 & 66.56 & 44.94 & 66.23 & 46.92 \\
				CB RW \cite{cui2019class}& 27.63 & 13.46 & 38.06 & 16.20 & 66.01 & 42.88 & 78.69 & 47.52 \\
				CB Focal \cite{cui2019class} & 25.43 & 12.90 & 39.73 & 16.54 & 63.98 & 42.01 & 80.24 & 49.98 \\ \hline
				HG-DRS \cite{cao2019learning} & 27.16  & 14.03 & 29.93 & 14.85 & - & - & -   & -     \\
				LDAM-HG-DRS \cite{cao2019learning} & 24.42  & 12.72 & 24.53 & 12.82 & - & - & -   & -     \\
				M-DRW \cite{cao2019learning} & 24.94  & 13.57 & 27.67 & 13.17 & 59.49 & 43.78 & 58.91 & 44.72 \\
				LDAM-DRW \cite{cao2019learning} & 22.97  & 11.84 & 23.08 & 12.19 & 57.96 & 41.29 & 54.64 & 40.54 \\ 
				LDAM-DRW + SSP \cite{yang2020rethinking}  & 22.17  & 11.47 & 22.95 & 11.83 & 56.57 & 41.09 & 54.28 & 40.33 \\ \hline
				\textbf{MM-DRW} (ours) & \textbf{21.98} & \textbf{11.44} & \textbf{22.83} & \textbf{11.48} & \textbf{57.14} & \textbf{40.63} & \textbf{54.57} & \textbf{40.28}  \\ 
				\textbf{MM-LDAM-DRW} (ours) & \textbf{21.37} & \textbf{11.26} & \textbf{21.82} & \textbf{11.33} & \textbf{56.53} & \textbf{40.54} &\textbf{53.70} & \textbf{40.07}  \\ \hlinewd{1.pt}
			\end{tabular}
		\end{adjustbox}
		\label{table:top-1}
	\end{table*}
	
	\noindent
	\textbf{Baselines.} We compare our MM with the standard training and other state-of-the-art algorithms. For a fair compararision, we follow the prior experiment setting \cite{cao2019learning}: (1) Empirical risk minimization (ERM) loss: all the examples have the same weights; by default, all model use standard cross-entropy loss; (2) Re-Weighting (RW): the model re-weights each sample by the inverse of the sample size of its class, and then re-normalize to make the weights $1$ on average in the mini-batch; (3) Re-Sampling (RS): each example is sampled with probability proportional to the inverse sample size of its class; (4) CB : the examples are re-weighted or re-sampled according to the inverse of the effective number of samples in each class, defined as (1-$\beta^{n_i}$) = (1-$\beta$), instead of inverse class frequencies; (5) Focal: we use the recently proposed focal loss; (6) SGD schedule: by SGD, we also refer to the standard schedule where the learning rates are decayed a constant factor at certain steps; we follow the same standard learning rate decay schedule. 
	
	\noindent
	\textbf{Our proposed algorithms} We evaluate the following algorithms: (1) MM : the proposed Maximum Margin losses; (2) MM-DRW : following the training Algorithm \ref{al:effective}, the MM with DRW Eq.\ref{eq:delta} is evaluated and (3) MM-LDAM-DRW with Eq. \ref{eq:delta_pos} and \ref{eq:delta_neg} is also performed with the parameter settings in Table. \ref{ab:cifar10} and \ref{ab:cifar100}.
	
	\noindent
	\textbf{Implementation details for CIFAR.} For CIFAR-$10$ and CIFAR-$100$, we follow the simple data augmentation in \cite{he2016deep} for training: $4$ pixels are padded on each side, and a $32 \times 32$ crop is randomly sampled from the padded image or its horizontal flip. We use ResNet-$32$ \cite{he2016deep} as our base network, and use stochastic gradient descend with momentum of $0.9$, weight decay of $2 \times 10^{-4}$ for training. The model is trained with a batch size of $128$ for $200$ epochs. For a fair comparison, we use an initial learning rate of $0.1$, then decay by $0.01$ at the $160$th epoch and again at the $180$th epoch. We also use linear warm-up learning rate schedule \cite{goyal2017accurate} for the first $5$ epochs. 
	
	\subsection{Experimental results on CIFAR}
	\textbf{Imbalanced CIFAR-10 and CIFAR-100.} The original version of CIFAR-10 and CIFAR-100 contains $50,000$ training images and $10,000$ validation images of size $32 \times 32$ with $10$ and $100$ classes, respectively. We evaluate the MM loss function on their imbalanced version that reduces the number of training examples per class and keeps the validation set unchanged. To ensure that our methods are compared with a variety of settings, we consider two types of imbalance: long-tailed imbalance \cite{cui2019class} and step imbalance \cite{buda2018systematic}. We use imbalance ratio $\rho$ to denote the ratio between sample sizes of the most frequent and least frequent class, i.e., $\rho=\max_i\{n_i\}/\min_i\{n_i\}$. A long-tailed imbalance follows an exponential decay in sample sizes across different classes. For step imbalance setting, all minority classes have the same sample size, as do all frequent classes. This gives a clear distinction between minority classes and majority classes.
	
	\noindent
	\textbf{Performances.} We report the top-$1$ validation error of various methods for imbalanced versions of CIFAR-$10$ and CIFAR-$100$ in Table \ref{table:top-1}. We evaluate the performances of MM as well as MM with DRW training schedule. The overall performance of MM is better than LDAM. The MM-DRW also shows the effectiveness, compared with LDAM-DRW. The combination of MM and LDAM with DRW represents the best performances in this experimental setting. To show the effectiveness, we only compare the results with \cite{cao2019learning}.
		
	\begin{figure}
	\centering
	\includegraphics[width=0.98\columnwidth]{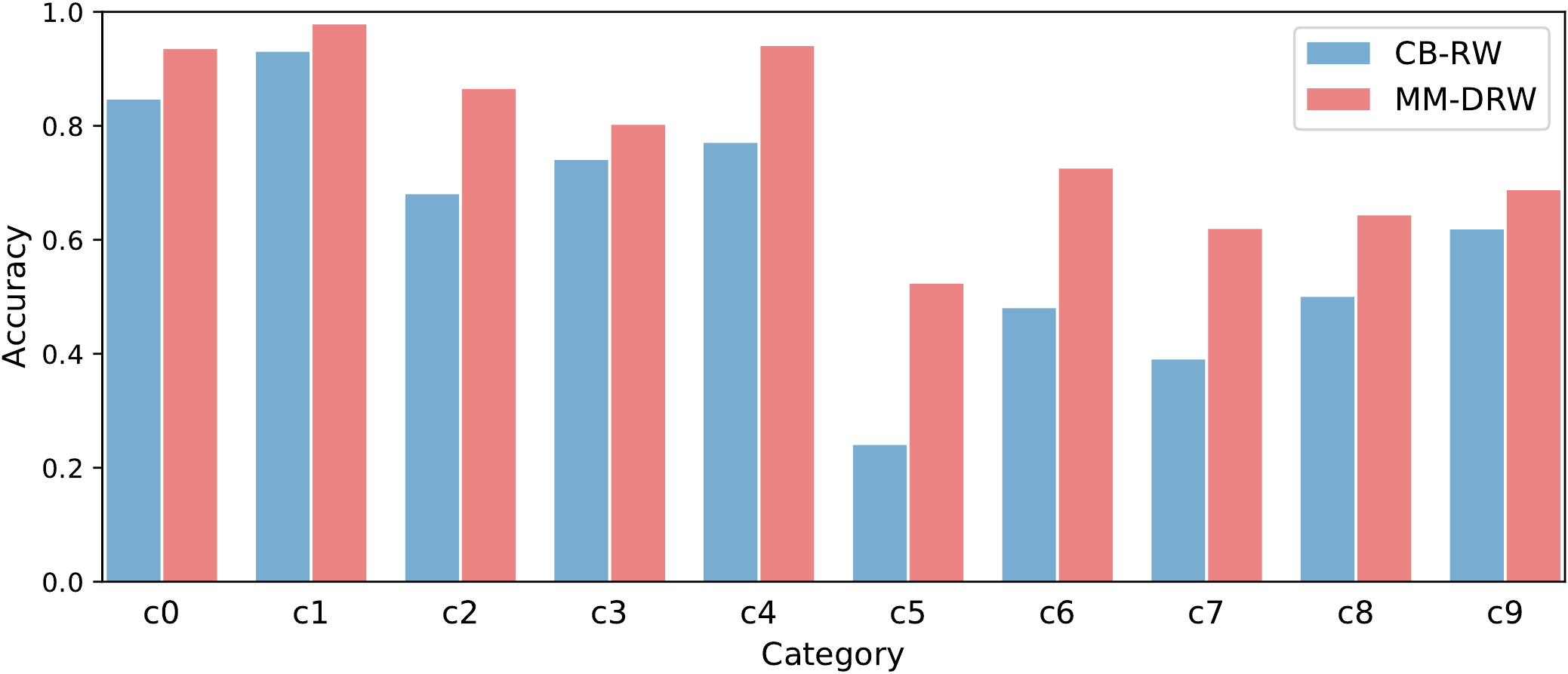}
	\caption{Per-class top-1 error on CIFAR-10 with step imbalance ($ \rho= 100$). Classes $C1$ to $C4$ are majority classes, and the rest are minority classes. Under this extremely imbalanced setting, CB-RW suffers from under-fitting while the proposed MM-DRW exhibits better generalization on the overall classes.}
	\label{fig:acc_cifar10}
	\end{figure}

	\subsection{Ablation study}	
	\textbf{Generalization.} To show the effectiveness of the MM loss function, we show the per-class error of CB-RW  in Figure \ref{fig:acc_cifar10} on the imbalanced CIFAR-$10$.
	
	\noindent
	\textbf{Hyper-parameters.} To show how to achieve the performances of MM-DRW, we prepare the ablation study on the hyper-parameters $\delta^{+}, \delta^{-}$ in Eq. \ref{eq:hmm_pos} and \ref{eq:hmm_neg}, as shown in Table. \ref{ab:cifar10} and \ref{ab:cifar100}, respectively. The results show that the ratio of $\delta^{+}/\delta^{-} > 1.0$ depends on label distributions (long-tailed or step) and dataset size.  
	
\begin{table}[ht]
	\centering
	\caption{Ablation Study : top-1 validation errors of hyper-parameters $\delta^{+} =\delta^{-} * \beta $ (Eq. \ref{eq:hmm_pos} and \ref{eq:hmm_neg})  on CIFAR-10.}
	\begin{adjustbox}{width=0.48\textwidth}
	\begin{tabular}{l|ll|ll|ll|ll}
		\hlinewd{1.1pt}
		Dataset  & \multicolumn{8}{l}{Imbalanced CIFAR-10}                                                                                        \\ \hline
		Type     & \multicolumn{2}{l|}{long-tailed} & \multicolumn{2}{l|}{step}    & \multicolumn{2}{l|}{long-tailed} & \multicolumn{2}{l}{step}  \\ \hline
		Ratio    & \multicolumn{1}{l|}{100}   & $\beta / \delta^{-}$   & \multicolumn{1}{l|}{100} & $\beta / \delta^{-}$ & \multicolumn{1}{l|}{10}    & $\beta / \delta^{-}$   & \multicolumn{1}{l|}{10} & $\beta / \delta^{-} $ \\ \hline
		\multirow{4}{*}{MM-DRW} & 22.24 &  1.4 / 0.6 & 22.92                & 1.2 / 0.6  & 11.66              & 1.1 / 0.7  &  \textbf{11.48}  & 1.0 / 2.1  \\
		                                 & \textbf{21.98}  &  1.5 / 0.6 & \textbf{22.83}  & 1.3 / 0.6  & \textbf{11.44} & 1.2 / 0.7  &11.64                 & 1.1 / 2.1    \\
		                                     &  22.43         &  1.6 / 0.6  &  23.29             & 1.4 / 0.6  & 11.86              & 1.3 / 0.7 & 11.78                 & 1.2 / 2.1 \\ \hlinewd{1.1pt}
	\end{tabular}
}
\end{adjustbox}
\label{ab:cifar10}
\end{table}

\begin{table}[ht]
	\centering
	\caption{Ablation Study : top-1 validation errors of hyper-parameters $\delta^{+} =\delta^{-} * \beta $ (Eq. \ref{eq:hmm_pos} and \ref{eq:hmm_neg}) on CIFAR-100.}
	\begin{adjustbox}{width=0.48\textwidth}
	\begin{tabular}{l|ll|ll|ll|ll}
		\hlinewd{1.1pt}
		Dataset  & \multicolumn{8}{l}{Imbalanced CIFAR-100}                                                                                        \\ \hline
		Type     & \multicolumn{2}{l|}{long-tailed} & \multicolumn{2}{l|}{step}    & \multicolumn{2}{l|}{long-tailed} & \multicolumn{2}{l}{step}   \\ \hline
		Ratio    & \multicolumn{1}{l|}{100}  & $\beta / \delta^{-}$   & \multicolumn{1}{l|}{100} & $\beta / \delta^{-}$ & \multicolumn{1}{l|}{10}    & $\beta / \delta^{-}$   & \multicolumn{1}{l|}{10} & $\beta / \delta^{-} $ \\ \hline
		\multirow{4}{*}{MM-DRW} & 58.02  & 1.2 / 1.2  & 54.65              & 1.7 / 1.8  & 40.97              & 1.3 / 1.5 & 40.42               &  1.0 / 2.4  \\
								   & \textbf{57.14}     & 1.3 / 1.2  & \textbf{54.57} & 1.8 / 1.8  & \textbf{40.63} & 1.4 / 1.5 & \textbf{40.28}  &  1.1 / 2.4  \\
								           & 57.24         & 1.4 / 1.2  &  54.76              & 1.9 / 1.8  &  40.95             & 1.5 / 1.5 & 40.48              &  1.2 / 2.4  \\ \hlinewd{1.1pt}
	\end{tabular}
	\end{adjustbox}
	\label{ab:cifar100}
\end{table}
	
	\section{Conclusion}\label{sec:conc}
	The Maximum Margin (MM) was proposed for considering the class-imbalance data learning issue: the trained model tends to predict the majority classes rather than the minority ones. That is, overfitting for minority classes seems to be one of the challenges of generalization. For a good generalization on the minority classes, we designed the Maximum Margin (MM) loss function, motivated by minimizing a margin-based generalization bound through the shifting decision bound. The theoretically-principled label-distribution-aware margin (LDAM) loss was successfully applied with prior strategies such as re-weighting or re-sampling, along with the training schedule. However, they didn't investigate the maximum margin loss led by samples yet. In this study, we showed the effectiveness of the proposed maximum margin with LDAM's training schedule on artificially imbalanced CIFAR-10/100.

	\newpage
	
	\bibliographystyle{IEEEbib}
	\bibliography{rib}
\end{document}